\documentclass[10pt,twocolumn,letterpaper]{article}

\usepackage{cvpr}              %

\usepackage{graphicx} 
\usepackage{amsmath}
\usepackage{booktabs}
\usepackage{xspace} %

\usepackage{algorithm}
\usepackage{algorithmic}
\usepackage[utf8]{inputenc} %
\usepackage[T1]{fontenc}    %
\usepackage[dvipsnames]{xcolor}

\newcommand{\RNum}[1]{\uppercase\expandafter{\romannumeral #1\relax}}

\newcommand{\ourmethod}{\texttt{RANLEN}\xspace}

\title{Local Low-light Image Enhancement via Region-Aware Normalization}
\definecolor{cvprblue}{rgb}{0.21,0.49,0.74}
\usepackage[pagebackref,breaklinks,colorlinks,citecolor=cvprblue]{hyperref}

\def\confName{CVPR}
\def\confYear{2024}

\author{Shihurong Yao$ ^{1} $ \qquad
Yizhan Huang$ ^{1} $ \qquad
Xiaogang Xu$ ^{2,3} $ \qquad
\\
$ ^{1} $Department of Computer Science and Engineering, The Chinese University of Hong Kong
\\
$ ^{2} $Zhejiang University $  \qquad
^{3} $Zhejiang Lab
\\
{\tt\small \{shryao1, yzhuang22\}@cse.cuhk.edu.hk, xiaogangxu00@gmail.com}
}
\begin{document}
\maketitle
\begin{abstract}

In the realm of Low-Light Image Enhancement (LLIE), existing research primarily focuses on enhancing images globally. However, many applications require local LLIE, where users are allowed to illuminate specific regions using an input mask, such as creating a protagonist stage or spotlight effect. However, this task has received limited attention currently. This paper aims to systematically define the requirements of local LLIE and proposes a novel strategy to convert current existing global LLIE methods into local versions.  The image space is divided into three regions: Masked Area A be enlightened to achieve the desired lighting effects; Transition Area B is a smooth transition from the enlightened area (Area A) to the unchanged region (Area C). To achieve the task of local LLIE, we introduce Region-Aware Normalization for Local Enhancement, dubbed as \ourmethod. \ourmethod   
uses a dynamically designed mask-based normalization operation, which enhances an image in a spatially varying manner, ensuring that the enhancement results are consistent with the requirements specified by the input mask. Additionally, a set of region-aware loss terms is formulated to facilitate the learning of the local LLIE framework. Our strategy can be applied to existing global LLIE networks with varying structures. Extensive experiments demonstrate that our approach can produce the desired lighting effects compared to global LLIE, all the while offering controllable local enhancement with various mask shapes.
\end{abstract}

\begin{figure}[ht]
  \includegraphics[width=\columnwidth]{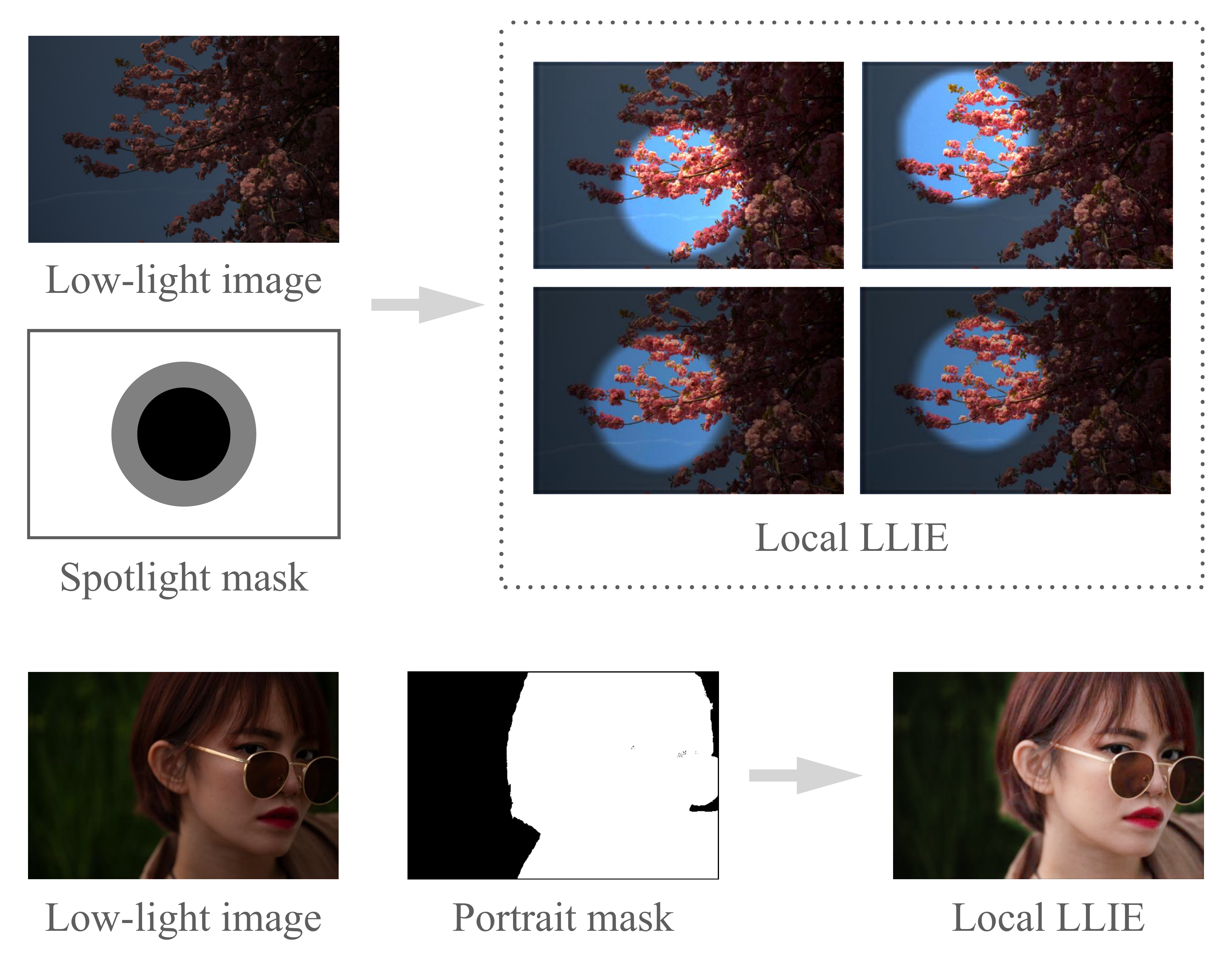}
  \vspace{-0.3in}
  \caption{With different mask, \ourmethod enables image enhancement within local areas and with different degree of enlightening.}
  \vspace{-0.2in}
  \label{fig:teaser}
\end{figure}

\section{Introduction}
\label{intro}

With the popularization of readily-available cameras on various devices, photo sharing has become very common nowadays.
However, these captured photos can be underexposed due to low- and back-lighting conditions, which are unsuitable for propagation via the internet.
Therefore, the technique of low-light image enhancement (LLIE) has been proposed in recent years~\cite{sid,DeepUPE,2020MIRNet,star}, which refers to the task of enhancing an image taken in a low-light environment to a normally-exposed one.
The traditional LLIE techniques, which are not learning-based, have been integrated into commercial softwares, e.g., Adobe Photoshop.
However, using these softwares requires professional experiences and human efforts.
With the development of deep learning, many learning-based LLIE approaches have been proposed to complete the enhancement automatically.
All current learning-based LLIE methods focus on \textit{global} image enhancement and can not implement the local modification.
Nevertheless, there is also need of \textit{local} image enhancement in the real-world scenario. Taking portraits as an example, for images taken in underexposed environments, or in complex lighting cases, various regions should be enhanced differently according to the user requirement.

In this work, we introduce a new task setting, Local LLIE, where users can complete the local enhancement for an underexposed image with an input binary mask map. 
The mask indicates the specific area to be enhanced, and there shall be a smooth transition from the enlightened area to the unenlightened region.
There are no existing strategies that can solve the Local LLIE task.

To solve the above task, a key question is: given a global enhancement network $g(\cdot)$, how can we encode the information of the mask without significant modification to its backbone, and meanwhile keep the corresponding enhancement effects in the target region? A simple solution would be modifying the input to the concatenation of the input image and the mask map.  However, merely encoding the mask information in the input stage can not guarantee the mask information expressed in the output. This is due to the potential loss of mask data at various levels within the network's backbone.

On the other hand, we propose \ourmethod to encode the mask information into hierarchical levels of the backbone in $g(\cdot)$. Specifically, in selected layers of $g(\cdot)$, we feed the mask map into a learnable neural network to predict the parameters of spatially-adaptive normalization operations, and apply the normalization. To address the challenge of achieving smooth transition between the enhanced and the unchanged area, we propose a training scheme that first divides the input image into three different areas, i.e., area \textit{A}, \textit{B}, and \textit{C}.
Area $A$ is the area to enhance; Area $B$ is the area for transition; Area $C$ is the area not required to be enlightened.
Then, we propose to apply different loss functions on these distinct spatial regions to achieve the corresponding purposes.
For area $A$, we keep the loss applied on the original model $g$ since this area should be enlightened to the normal light.
In area $B$, we employ a novel gradient-smooth loss to minimize the second-order gradient of the illumination component.
In area $C$, we implement a reconstruction loss between the predicted image and input image, since $C$ is not required to be changed.
Such a training scheme is general for existing LLIE approaches with various structures. 
Thus, \ourmethod can be easily integrated into existing global enhancement networks $g(\cdot)$ with minor modifications. 
Moreover, combining \ourmethod with the Retinex-based models like DeepUPE~\cite{DeepUPE}, LIME~\cite{lime}, we may achieve both local editing and degree control. This further improves the flexibility for software users, as shown in Fig.~\ref{fig:teaser}.

To demonstrate the effectiveness of the proposed method, we conduct extensive experiments on two public datasets with the backbone of $g(\cdot)$ as a Convolutional Neutral Network (CNN), e.g.,  DeepUPE~\cite{DeepUPE}, or Transformer, e.g., STAR~\cite{star}.
The experimental results demonstrate that for a low-light image, the proposed framework can enlighten the specified local area to different degrees, as displayed in Fig.~\ref{fig:teaser}. Moreover, it is suitable for current approaches with different parametric outputs and both convolutional and transformer-based structures.

In summary, our contribution is three-fold.
\begin{enumerate}
    \item We introduce a novel task setting, Local LLIE, which is to enhance an input image given a binary mask, as is shown in Fig.~\ref{fig:teaser}, satisfying the requirement of local image enhancement. 
    \item For Local LLIE, we propose a plug-and-play method \ourmethod,  to convert existing networks that are originally designed for global editing into local ones. Moreover, we propose a novel training scheme, applying different loss components in spatial-varying areas. In particular, a novel gradient-smooth loss is proposed to achieve the smooth transition effect.
    \item We conduct extensive experiments on public datasets with representative model structures, showing that the proposed approach and the corresponding training scheme can meet the requirements of Local LLIE.
\end{enumerate}

\section{Related Work}
To the best of our knowledge, there is no method working under the local low-light image enhancement setting. The most relevant work is ABPN~\cite{Lei_2022_CVPR}, which proposes an adaptive blend pyramid network for local photo retouching. Their method is verified under two tasks: cloth retouching and face retouching. The most significant difference is that, our method supports explicit mask input, while ABPN doesn't.

In this section, We briefly introduce the recent CNN-based and Transformer-based ``Global Underexposed Photo Enhancement models", and then introduce some LLIE methods containing spatially-varying operations. However, note that their learning objectives and evaluation metrics still focus on the enhanced \textit{global} result, which is different from our \textit{local} learning objectives.

\begin{figure*}[t]
        \centering
        \includegraphics[width=\textwidth]{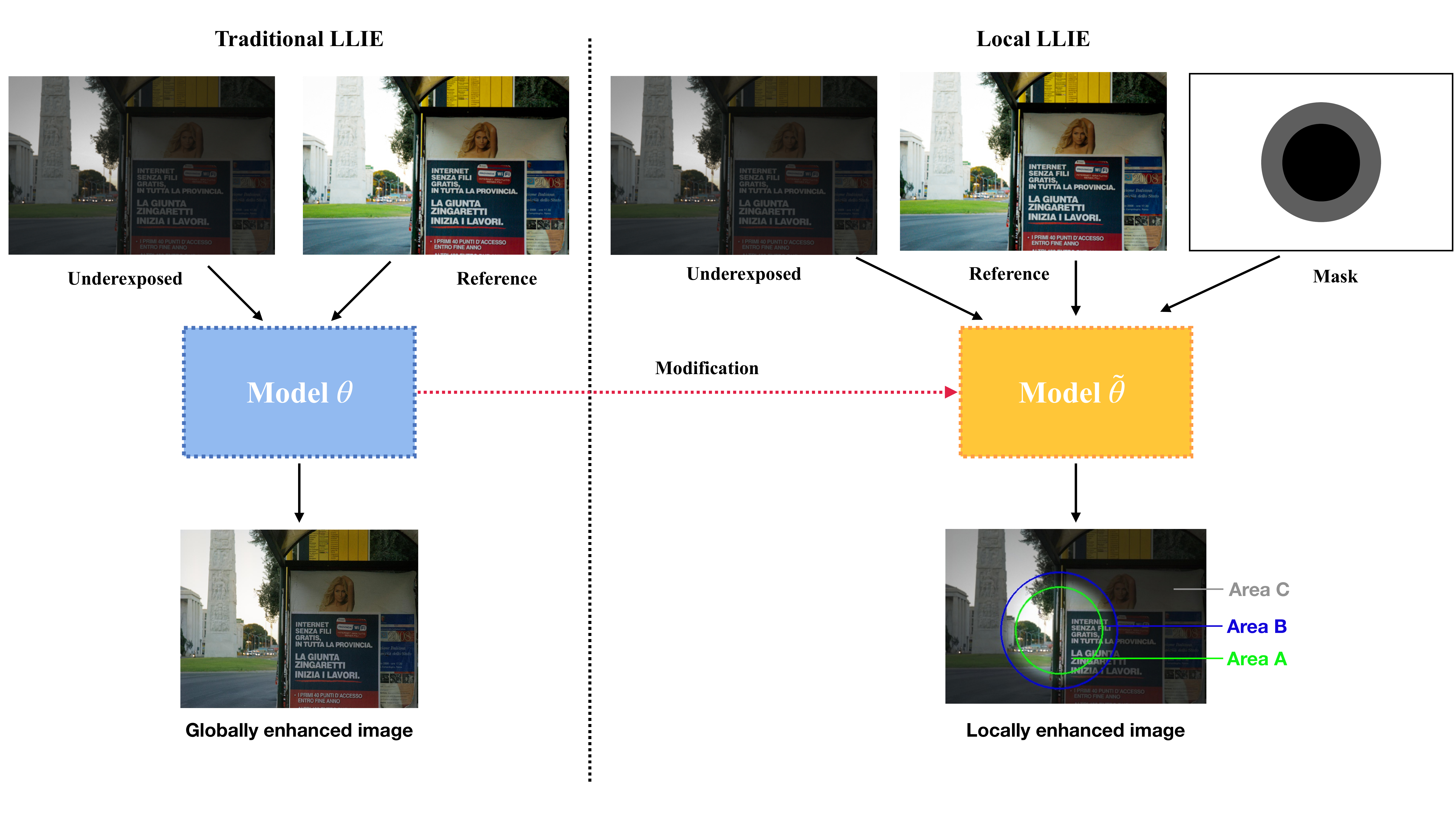}
        \vspace{-0.3in}
        \caption{Comparison of traditional LLIE setting and our new Local LLIE setting. The traditional LLIE method enhances an underexposed image across all the spatial areas, whereas the local LLIE task requires enhancements on a selective spatial region, indicated by a mask specified by the users. This results in a more tailored enhancement in the local LLIE, addressing varying levels of exposure within different parts of the image.}
        \vspace{-0.2in}
        \label{setting}
\end{figure*}

\subsection{Global Low-light Image Enhancement}
LLIE is a vital while challenging task,  since the underexposed areas are usually imperceptible, and the enhancement process is highly nonlinear.
With the massive success of deep learning, a significant trend for solving this task is the CNN-based image enhancement method, and several approaches have achieved promising results. 
Another new trend is the Transformer-based models, starting from Vision Transformer~\cite{vit}, which brings Transformers~\cite{transformer} to the field of Computer Vision.

CNN models mostly use an encoder, followed by an output tail. In general, the output tail could be divided into several categories:
\begin{itemize}
    \item Retinex-theory-based methods. These models use Retinex Theory to predict the illumination map, hence generating the reflectance image. Examples are LIME~\cite{lime} and URetinexNet~\cite{uretinexnet}, and Retinexformer~\cite{Cai_2023_ICCV}.
    \item Parameterized models. These models predict parameters of predefined filters, e.g., DeepLPF~\cite{deeplpf}, or enhancement curves, e.g., ZeroDCE~\cite{Zero-DCE}.
    \item Encoder-Decoder-based models. These kinds of models are prevalent via predicting the pixel value directly, e.g., SID~\cite{sid}, MIRNet~\cite{2020MIRNet}.
    \item 3D-Look-up-Table-based approaches. Such strategies predict the index to retrieve the pixel value from 3D look-up table, like~\cite{3dlut1} and~\cite{3dlut2}.
    \item Diffusion-based models. As a powerful technique in generative modeling, diffusion models have received a huge amount of attention recently. There are research works that adapt a general diffusion model to the setting of LLIE. Examples include Diff-Retinex~\cite{diff-retinex}, and ExposureDiffusion~\cite{wang2023exposurediffusion}.
\end{itemize}

Another popular direction is Transformer-based models. ViT~\cite{vit} bridges the gap of Transformer between Natural Language Processing (NLP) and Computer Vision (CV). Inspired by ViT, much Transformer-based work towards low-level CV tasks appears. With over 114 Million parameters and 33 GFLOPS, IPT~\cite{ipt} model introduces contrastive learning to ViT, adapting to different image processing tasks, e.g., super-resolution, denoising, deraining. After that, STAR~\cite{star} reduces model complexity to a large margin, and still improves the performance, benefiting from its Long-short Range Transformer Module.

All these models mentioned above enhance the global area, and the corresponding performances are evaluated globally. Thus, they are not capable of performing local enhancement.

\subsection{LLIE with Spatially-varying Operations}
Tian and Cohen~\cite{tian2017} proposed a method using global and local enhancement for color enhancement. The algorithm includes both global contrast adaptive enhancement and a hue-preserving local contrast adaptive enhancement module. Finally, a contrast-brightness-based fusion algorithm obtains the final result, which represents a trade-off between global contrast and local contrast. However, such a method can not enhance local areas according to user-given inputs.

Moran et al.~\cite{deeplpf} proposed Deep Local Parametric Filters to learn spatially local filters of three different types (Elliptical Filter, Graduated Filter, Polynomial Filter). CNNs are introduced to regress the parameters of these spatially localized filters. Nevertheless, these local filters are still designed for global enhancement.

DeepExposure~\cite{deepexposure} is an algorithm for learning local exposures with deep reinforcement adversarial learning. DeepExposure leverages image segmentation to segment the input image into sub-images based on low-level features. Then, the algorithm learns to enhance with different local exposure increments for each sub-image, i.e., a local exposure for each sub-image, and fuse the sub-images to get the final enhanced image. However, this approach can not enhance different image regions according to the user-given inputs.

\section{Method}
\subsection{Tasking Setting}
We introduce a new task setting, namely Local LLIE.  Local LLIE algorithms shall take two inputs: image $I\in \mathbb{R}^{n\times m\times 3} $, and a binary mask $M\in \mathbb{R}^{n\times m\times 2}$, where $M$ is specified by the user.
The task is to seek a learned enhancement map $f_\theta(I)$, that enhances the image in the masked area, and transits naturally to the dark area. Note that we denote the area to be enhanced with the mask value of 1 and the area to be kept with 0.

Under the Local LLIE setting, an input image can be divided into Masked Area $A$ (encoded by the first channel of Mask $M$), Transition Area $B$ (encoded by the second channel of the mask $M$), and Unmasked Area $C$ (implicitly encoded by the mask $M$). 
Their details are explained as the following.
\label{areas}
\begin{enumerate}
    \item Masked Area $A$ is the region to be enlightened, which is specified by the first channel of $M$. And in the second channel, we also set the Area $A$ as 1. Formally speaking, $A=\{I_{i,j,k}|M_{i,j,0}=M_{i,j,1}=1,1\leq i\leq m, 1\leq j\leq n, k\in \{0,1,2\}\}$\footnote{Here, we refer the first channel of image as channel 0.}.
    \item Transition Area $B$ is the region that is supposed to achieve a smooth transition from Area $A$ to Area $C$, specified by the second channel of $M$. Formally speaking, $B=\{I_{i,j,k}|M_{i,j,0}=0, M_{i,j,1}=1,1\leq i\leq m, 1\leq j\leq n, k\in \{0,1,2\}\}$.
    \item Unmasked Area $C$ is the region to keep dark. It is the region that is not covered by any channel of the mask. Formally speaking, $C=\{I_{i,j,k}|M_{i,j,1}=M_{i,j,0}=0,1\leq i\leq m, 1\leq j\leq n, k\in \{0,1,2\}\}$.
\end{enumerate}

Fig.~\ref{setting} further illustrates the task setting of Local LLIE.   We designate the learning objective $L$ such that Area $A$ is to be enhanced effectively,  Area $B$ transits smoothly, and Area $C$ is to be kept dark. Details of loss functions will be described in the section for loss function. Under the Local LLIE setting, this work studies a model conversion problem: given an existing global image enhancement model $g(\cdot)$ with the parameters of $\theta$, we want a new model $f(\cdot)$, whose parameters $\tilde{\theta}$ are modified from $\theta$ with minimal efforts, to achieve local image enhancement. 
$\tilde{\theta}$ learns the transformation $f_{\tilde{\theta}}$ such that the learning objective $L(\hat{Y},Y)$ is minimized, where $\hat{Y}=f_{\tilde{\theta}}(I)$.

\subsection{Architecture Design of RANLEN}
\begin{figure}[htbp]
        \centering
        \includegraphics[width=\columnwidth]{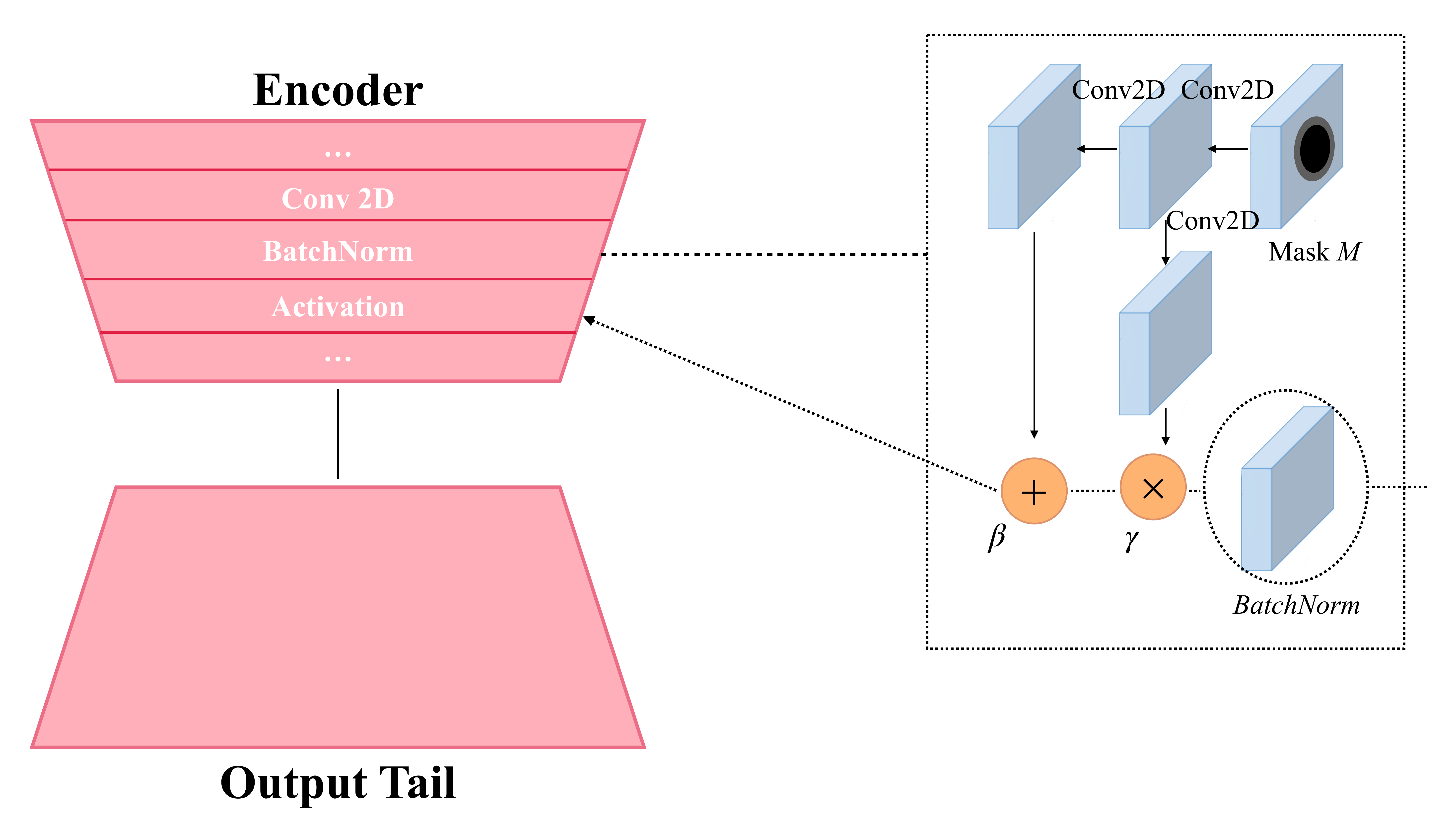}
        \caption{Architecture diagram of turning a global image enhancement model $g(\cdot)$ into local enhancement framework $f(\cdot)$. We select (a subset of) the normalization layers in the encoder part of $g(\cdot)$, and replace them with our \ourmethod Module. Inside the module, mask information is encoded via the convolution layer for spatially-varying enhancement.}
        \label{fig-arch}
\end{figure}

RANLEN simply replaces (some of) the normalization layers in $g_\theta$ by mask-encoded normalization layers. 
Unconditional normalization methods, e.g., Batch Normalization (BatchNorm)~\cite{batchnorm}, Instance Normalization (InstaceNorm)~\cite{instancenorm}, Layer Normalization (LayerNorm)~\cite{layernorm}, are not capable of encoding mask information.
In contrast, we propose to modulate activations in normalization layers through a spatially adaptive and learning-based transformation. 
The architecture of \ourmethod is illustrated in detail, as shown in  Fig.~\ref{fig-arch}.
To be specific,  a \ourmethod module first projects the mask into an embedding space and predicts the modulation parameters $\gamma$, $\beta$ via convolution. Then $\gamma$ and $\beta$ are multiplied and added to the normalized activation element-wise.

Formally speaking, in the context of Local LLIE, a \ourmethod module is as follows: for $k$-th convolution layer in a network $g_\theta$ with batch size $N$ with height $H^k$ and width $W^k$, let $h^k$ be the activation of $k$-th layer, $C^k$ be the number of channels in layer $k$. The mask-encoded activation value is:
\begin{equation}
\gamma^k_{c,y,x}(M)\frac{h^k_{n,c,y,x}-\mu_c^k}{\sigma^k_c}+\beta^k_{c,y,x}(M),
\end{equation}
where $h^i_{n,c,y,x} (n\in N, c\in C^k, y\in H^k, x\in W^k)$ is the activation before normalization, $\mu_c^k,\sigma_c^k$ are the mean and standard deviation of the activation in channel $c$. 
Here, $\gamma^i_{c,y,x}(M)$ and $\beta^k_{c,y,x}(M)$ are learning-based parameters converting the information of Mask $M$ to the scaling and bias values at the location $(c,y,x)$. 
$\gamma^i_{c,y,x}(M)$ and $\beta^k_{c,y,x}(M)$ are implemented with a simple two-layer CNN. Moreover, we add one more convolution layer after the output tail to refine the enhancement result.

\subsection{Loss Function Design}
\label{loss-func}
Given an image pair $(Y^i , \hat{Y}^i )$, we employ different loss functions in three areas, i.e., Masked Area $A$, Transition Area $B$, and Unmasked Area $C$.

We show the details of each loss component applied on different area in the following paragraphs, by taking DeepUPE as the example of $g_\theta$.

\noindent{\textbf{Masked area $A$.}}
For enlightening the Area $A$, we employ the original loss applied on the original model $g_\theta$. 
Note that different models may have varying loss choices for enhancement in loss component $L_A$. For instance, the loss function in DeepUPE is based on various constraints and priors on the illumination map, including $L_2$ Reconstruction Loss, Smoothness Loss, and Color Loss.

\begin{figure*}[t]
        \centering
        \includegraphics[width=\textwidth]{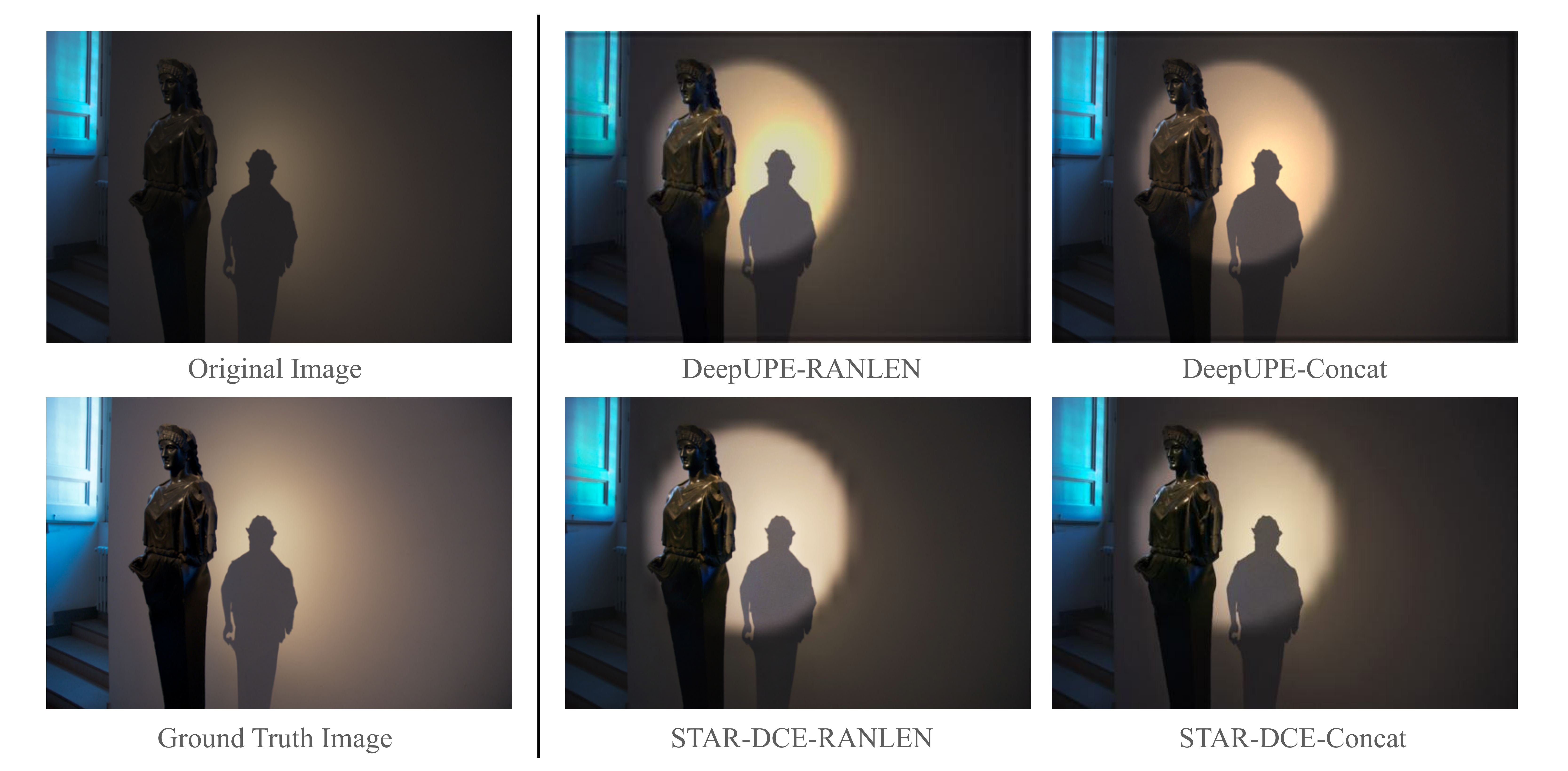}
        \vspace{-0.3in}
        \caption{Visual comparison between \ourmethod and \texttt{Concat} on FiveK dataset.}
        \vspace{-0.2in}
        \label{fivek-com}
\end{figure*}

\vspace{0.1in}
\noindent{\textbf{Transition Area $B$.}}
The transition Area $B$, between Masked Area $A$ and Unmasked Area $C$, plays a vital role in increasing the consistency of the different regions in the whole image. %
In this area, we should ensure a smooth transition from darkness to brightness.
Different from the smooth prior adopted in~\cite{smooth1,DeepUPE}, which smooth \textit{the first-order gradient} of the illumination map, the proposed Gradient-Smooth loss is working to smooth  \textit{the second-order gradient} of the illumination map. \footnote{For the non-Retinex-theory-based model, we directly apply the Gradient-Smooth Loss on the predicted image instead of the illumination map.}
In other words, we apply the smooth loss prior to the gradient of the illumination map $L(\cdot)$. Specifically, for each pixel $p$ in Area $B$, we can define the smooth loss $L_{B}^i$ as
\begin{align}
\begin{split}
L_{B}^{i}=\sum_p  &\omega_{x,x}^p (\partial_x \partial_x L(\hat{Y}^i))^2+ \omega_{y,x}^p (\partial_y \partial_x L(\hat{Y}^i))^2\\ 
    + & \omega_{x,y}^p (\partial_x \partial_y L(\hat{Y}^i))^2+\omega_{y,y}^p (\partial_y \partial_y L(\hat{Y}^i))^2,
\end{split}
\end{align}
\begin{equation}
    \omega^p_{u,v}=(|\partial_u \partial_v \log(I^i)^p|^s + \epsilon) ^{-1},  u,v\in \{x,y\},
\end{equation}
where $s=1.2$ controls the sensitivity to the second-order image gradient, and $\epsilon=0.0001$ is a small constant that prevents division by zero.

\vspace{0.1in}
\noindent{\textbf{Unmasked Area $C$.}}
For pixels in Area $C$, we want them to be consistent with \textit{the original dark image} $I$. In this case, we use the reconstruction error between the dark image $I$ and the predicted image $\hat{Y}$. Taking $L_2$ norm as an example,
the reconstruction loss can be written as
\begin{equation}
    L_{C}^{i}= ||(\hat{Y}^i- I^i)||_2.
\end{equation}

The total loss is designed to be the weighted combination of three loss components $L_A$, $L_B$, and $L_C$, as follows:
\begin{equation}
    L^i=\alpha L_A^i/r_a+\beta L_B^i/r_b + \gamma L_C^i/r_b, i\in [1, N],
\end{equation}
where $\alpha$, $\beta$, and $\gamma$ are loss weights, $r_a$, $r_b$, and $r_c$ are balancing parameters measuring the coverage  of Area $A$, $B$, and $C$, respectively.
\subsection{Interactive and Flexible Photo Enhancement}
With  \ourmethod, the image enhancement process could be more interactive and flexible.

\vspace{0.1in}
\noindent{\textbf{Flexible Mask Shape.}}
In general, arbitrary designs of the mask are feasible. The mask could be in regular shape, i.e., semantic-agnostic, like rectangular and circular. Or it could be semantic-aware and can even be user-generated strokes. 
For example, a mask for the human face enables face retouching. 
Meanwhile, we can use morphological erosion or dilation~\cite{mor} to the mask of Area $A$ to obtain Area $B$ without human efforts.
These flexible designs enable the users to enhance according to their own preferences.

For simplicity, we implement \textit{circular} masks for Area $A$ and $B$ in our experiment to directly compare our proposed model with baseline models. Details will be discussed in the experimental section.

\vspace{0.1in}
\noindent{\textbf{Control the degree of enlightening.}}
Low-light image enhancement is known to be an ill-posed task-- some enhancement software users may prefer darker images, while others prefer brighter ones. If we combine \ourmethod with the model $g_\theta$ that adopts curve-based or Retinex-based models for enhancement, we can control the degree of enlightening by multiplying a coefficient to the predicted curve/illumination map.
This makes the enhancement process more interactive, and provides the flexibility for software users to choose their own lighting based on their preferences.

\begin{figure*}[t]
        \centering
        \includegraphics[width=\textwidth]{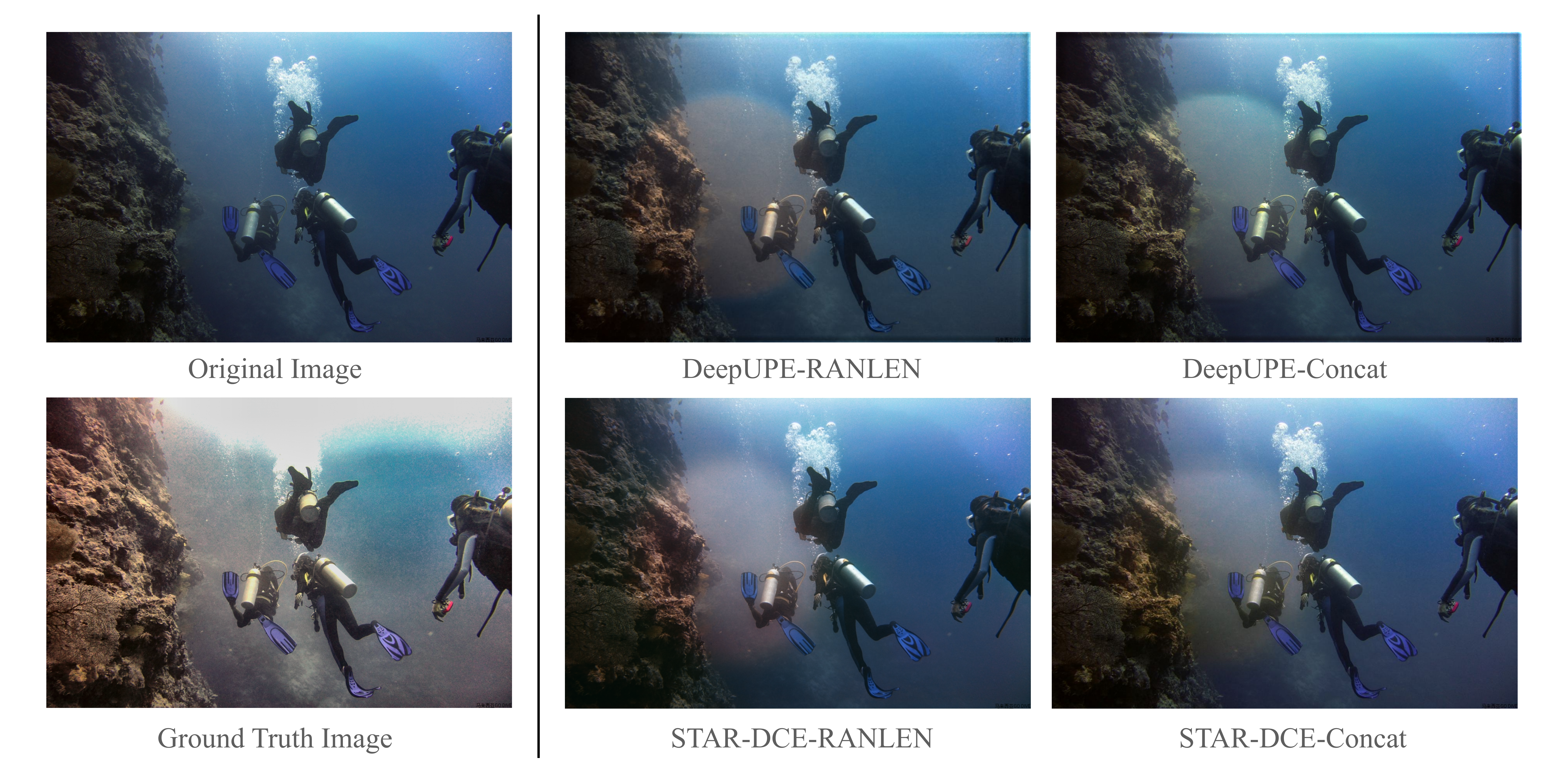}
        \vspace{-0.3in}
        \caption{Visual Comparison between \ourmethod and \texttt{Concat} on UIEB dataset.}
        \vspace{-0.2in}
        \label{uieb-com}
\end{figure*}

\section{Experiments}
\label{experiments}
\subsection{Experimental Setup}
\noindent\textbf{Dataset.} MIT-Adobe-FiveK~\cite{fivek} (FiveK) is used to evaluate our proposed  \ourmethod.  We follow the settings of previous methods~\cite{deeplpf,DeepUPE,lr} to only use the retouched images by Expert C. All images are resized to 256 pixels in the long edge, and the bit-depth is 8 bit. We select the first 4,500 images for training, and the remaining 500 images for validation and testing. 

Underwater Image Enhancement Benchmark (UIEB) is an underwater image enhancement dataset including 950 real-world underwater images, 890 of which have the corresponding reference images. Most of these underwater images suffer darkness, color casts, and blurring details.  We resize with 256 pixels in the long edge during training. We randomly select  812 paired images for training, and the remaining 78 paired images for validation and testing.

\vspace{1mm}
\noindent\textbf{Performance Metrics.} 
For Area $A$ of the enhanced image, we adopt two full-reference metrics to evaluate the image quality: PSNR (measured in dB), and SSIM~\cite{ssim} as our evaluation metrics in Area $A$. High PSNR and  SSIM suggest good image quality.  Note that both PSNR and SSIM are slightly adapted under the Local LLIE setting. For a mask $M$, an enhanced image $\hat{Y}$, and a reference image $Y$, the new PSNR and SSIM are measured between the image pair $M*\hat{Y}$ and $M*Y$, where $*$ is the element-wise product. Since the calculation is constrained within Area $A$, we normalize the PSNR result by multiplying a $\frac{1}{r_A}$ factor, where $r_A$ is the coverage of Area $A$ measured in percentage.

For Area $B$ of the enhanced image, so far, no proper metric can be used to quantitatively evaluate the transition effects. Here, we perform a qualitative analysis on the enhancement result of Area $B$.

\vspace{1mm}
\noindent\textbf{Baselines.}
We choose the method \texttt{Concat} as our baseline for comparison (as the straightforward approach mentioned in the introduction section). In \texttt{Concat}, we concatenate the mask input with the RGB channel of the input image.  In other words, the new model $f_{\tilde{\theta}}$ has the same structure as the global enhancement network $g_\theta$, except that we shall modify the model to accept a 5-channel input.

\vspace{1mm}
\noindent\textbf{Backbones.}
To comprehensively show the effectiveness of the proposed methods, we adopt \ourmethod and the \texttt{Concat} in both CNN-based and Transformer-based models. For CNN-based models, we choose DeepUPE as the representative one; for Transformer-based methods, we select STAR-DCE (STAR-based Zero-DCE backbone) as the typical one. Note that STAR and DeepUPE use different output tails -- DeepUPE is a Retinex-theory-based model, and STAR is a parameterized model. \footnote{Note that STAR and DeepUPE downsample the high-resolution images, process the image in a lower resolution, and upsample back to the original resolution. For these methods, we may need to adjust some parameters to avoid the upsampled image from generating ``blocking" visual effects.} Therefore, we construct four novel models, namely DeepUPE-\ourmethod, DeepUPE-\texttt{Concat}, STAR-DCE-\ourmethod and STAR-DCE-\texttt{Concat}. We both trained \ourmethod and \texttt{Concat} models for 300 epochs.

\subsection{Implementation Details}
Three loss component weights  $\alpha, \beta, \gamma$ are set to $1, 0.0001, 1$, respectively.   As for the implementation of the mask, we use circular masks for experiments. The circle in radius $r_1$ indicates Area $A$, and a larger circle in radius $r_2$ indicates Area $B$. In the following experiments, $r_1=2*m/7, r_2\sim U(1.2r_1,1.25r_1)$, where $m$ denotes the minimum between the image height and width. \par

For DeepUPE-\ourmethod, we add \ourmethod Module in the splat features extraction, global features extraction, local features extraction, and illumination mapping prediction part. 
For STAR-\ourmethod, we use the STAR-DCE backbone with the Mean Head tokenization Strategy. The network width is 32. We add a \ourmethod module in the tokenization. We keep the loss components in the original DeepUPE and STAR-DCE model and apply them in Area $A$ of the image.

\subsection{Experimental Results}
We show the enhanced images on the Adobe-MIT-FiveK dataset in Fig.~\ref{fivek-com}, and the enhanced images on the UIEB dataset in Fig.~\ref{uieb-com}. The output images are close to the normal-light ground-truth image in Area $A$, while keeping a smooth transition to the original input image.

\subsubsection{Quantitative Results in Area $A$}
PSNR and SSIM are measured only in Area \textit{A} of the enhanced images.
See Tab.~\ref{psnrssim-table} for PSNR and SSIM results. We can see that the results of RANLEN are better than the baseline.
The result shows \ourmethod both enjoy better visual qualities in Area A than \texttt{Concat}.
Moreover, we compute the global enhancement results and display them in Tab.~\ref{global-table}. As we can see, the local image enhancement network can also enjoy good image quality when compared to the global image enhancement network in Area A.

\begin{table}[htbp]
\centering
\begin{tabular}{lcc}
\hline
   Method Name             & UIEB & FiveK  \\\hline
STAR-DCE-\texttt{Concat} & 17.34 / 0.963   & 18.84 / 0.972   \\
STAR-DCE-\ourmethod  & \textbf{18.69} / \textbf{0.967}   &\textbf{19.03} / \textbf{0.974}    \\ \hline
DeepUPE-\texttt{Concat}  &  16.19 / 0.957  & 18.11 / 0.965  \\
DeepUPE-\ourmethod   & \textbf{18.03} / 0.957   & \textbf{18.33} / \textbf{0.967} \\
\hline
\end{tabular}
\caption{Image quality in Area \textit{A} measured in PSNR and SSIM for local enhancement networks.}
\label{psnrssim-table}
\end{table}

\begin{table}[htbp]
\centering
\begin{tabular}{lcc}
\hline
   Method Name               & UIEB & FiveK  \\\hline
STAR-DCE  & 19.00 / 0.972    &20.30 / 0.981    \\

DeepUPE & 14.83 / 0.944   & 18.27 / 0.963   \\\hline

\hline
\end{tabular}
\caption{Image quality in Area \textit{A} measured via PSNR and SSIM for global enhancement networks.}
\label{global-table}
\end{table}

\subsubsection{Ablation Study}

\begin{figure}[t]
        \includegraphics[width=\columnwidth]{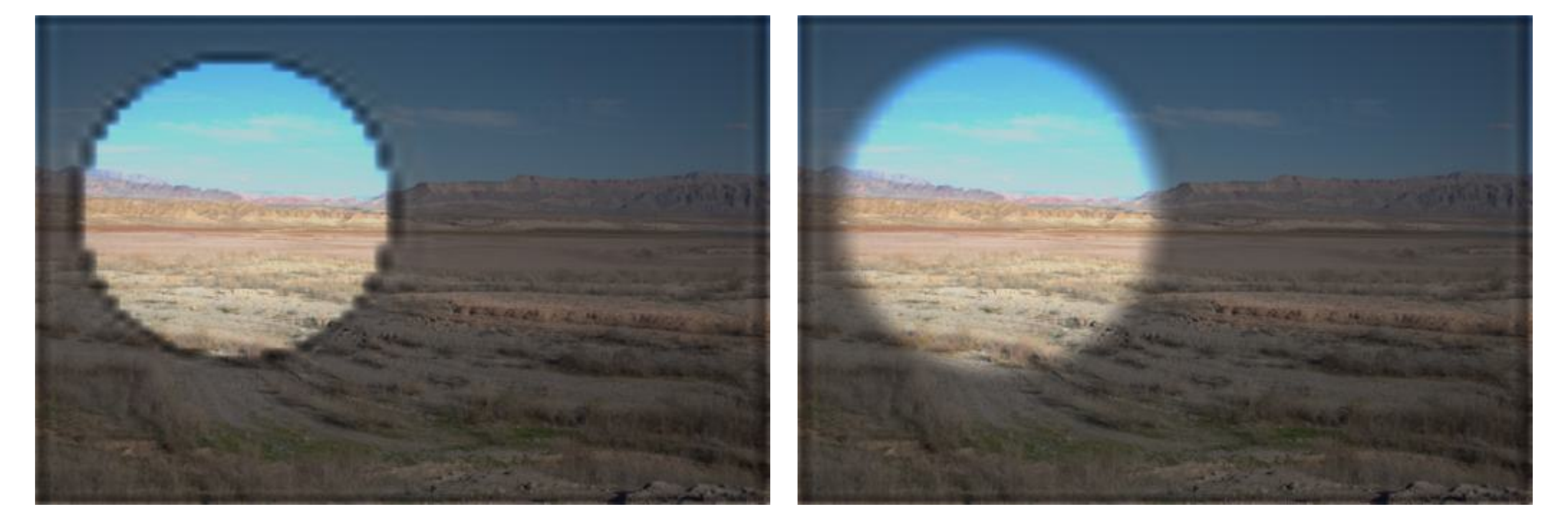}
        \caption{The ablation study for the gradient-smooth loss in DeepUPE. Left: models \textbf{without} gradient-smooth loss. Right: models \textbf{with} gradient-smooth loss.}
        \label{localllie-compare}
\end{figure}

\begin{figure}[t]
        \includegraphics[width=\columnwidth]{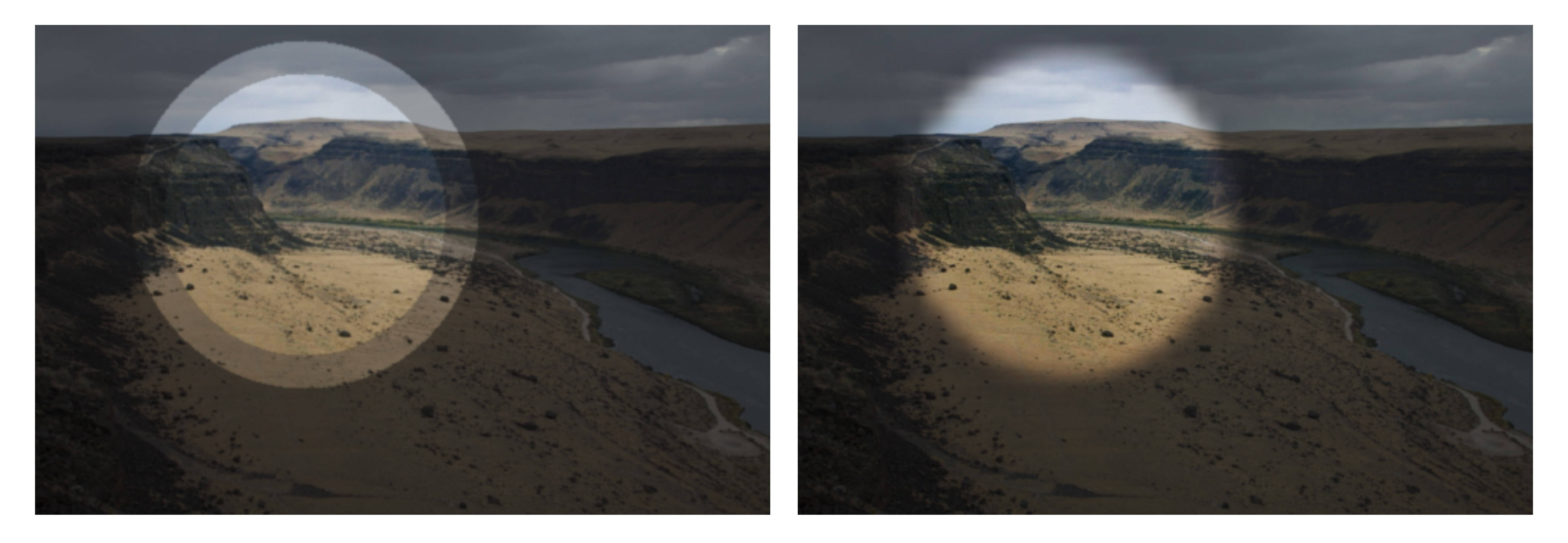}
        \caption{The ablation study for simply adding three tails for area decoding in STAR-DCE. Left: global enhancement STAR-DCE model with three decoder tails. Right: our proposed method STAR-DCE-\ourmethod.}
        \label{multitail}
\end{figure}

\paragraph{The gradient-smooth loss.}
The gradient-smooth loss is the key to ensure smooth transitions, which is also an innovation in our proposed training schemes. Thus, we illustrate the importance of gradient-smooth loss by removing the gradient-smooth loss. In other words, in our ablation experiment, the whole image is now divided into \textit{two} parts: Masked Area $A'=A$ and Unmasked Area $B'=\bar{A}$ (complement of Area $A$) specified by mask $M' \in \mathbb{R}^{n\times m \times 1} $, and we apply two loss components: Area $A$ loss $L_{A'}$ and reconstruction error $L_{B'}$ with weight $\alpha'=1, \beta'=1$. See Fig.~\ref{localllie-compare} for details. In general, for models trained without gradient-smooth loss, we spot sharp, unnatural edges in enhancement results.
\paragraph{Multi-cnn-and-concat in the tail.}
Another straightforward and intuitive way to come up with when computing local image enhancement is to use a global enhancement image network as the backbone and add three CNN tails as decoders to decode the information in three areas, respectively. The inputs to the first two tails are globally enhanced images, which are used to decode the image in Area A and Area B, and the input to the last tail is the original low-light image, which is used to interpret the information in Area C. After we get three outputs of the three tails, we can multiply by their area mask and add them together to get the final output. However, in Fig.~\ref{multitail} we can easily observe an unnatural transition between Area A and Area B, and between Area B and Area C.

\section{Applications}
This section demonstrates practical applications of \ourmethod.

\subsection{Control the Degree of Light}

\begin{figure}[t]
        \includegraphics[width=\columnwidth]{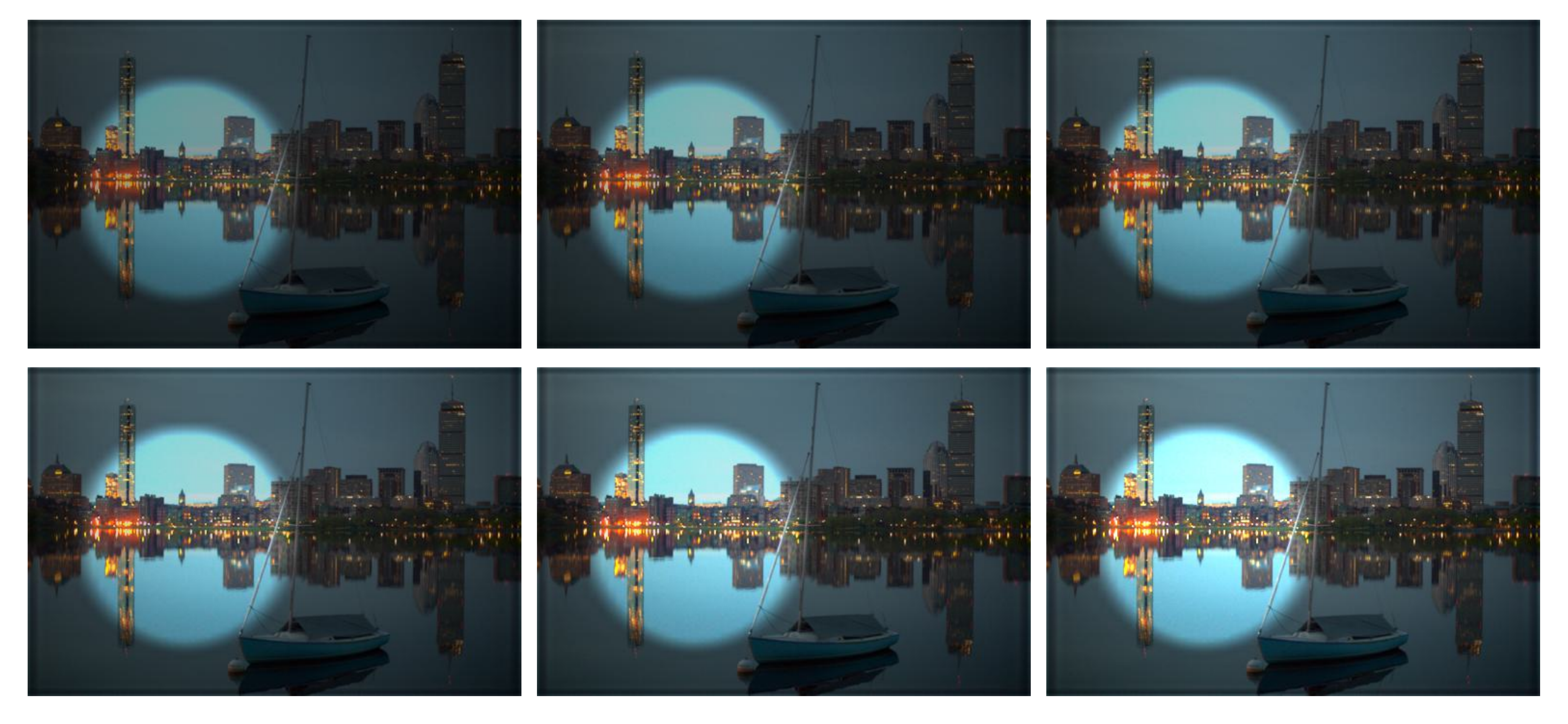}
        \caption{Control the enhancement results of DeepUPE-\ourmethod. The predicted illumination map is element-wise multiplied by a constant $\alpha$. The images get brighter with $\alpha$ decreasing. Left to right, up to down: $\alpha=1.3, 1.2, 1.1, 1.0, 0.9, 0.8$. }
        \label{localllie-degree}
\end{figure}

\begin{figure}[t]
        \includegraphics[width=\columnwidth]{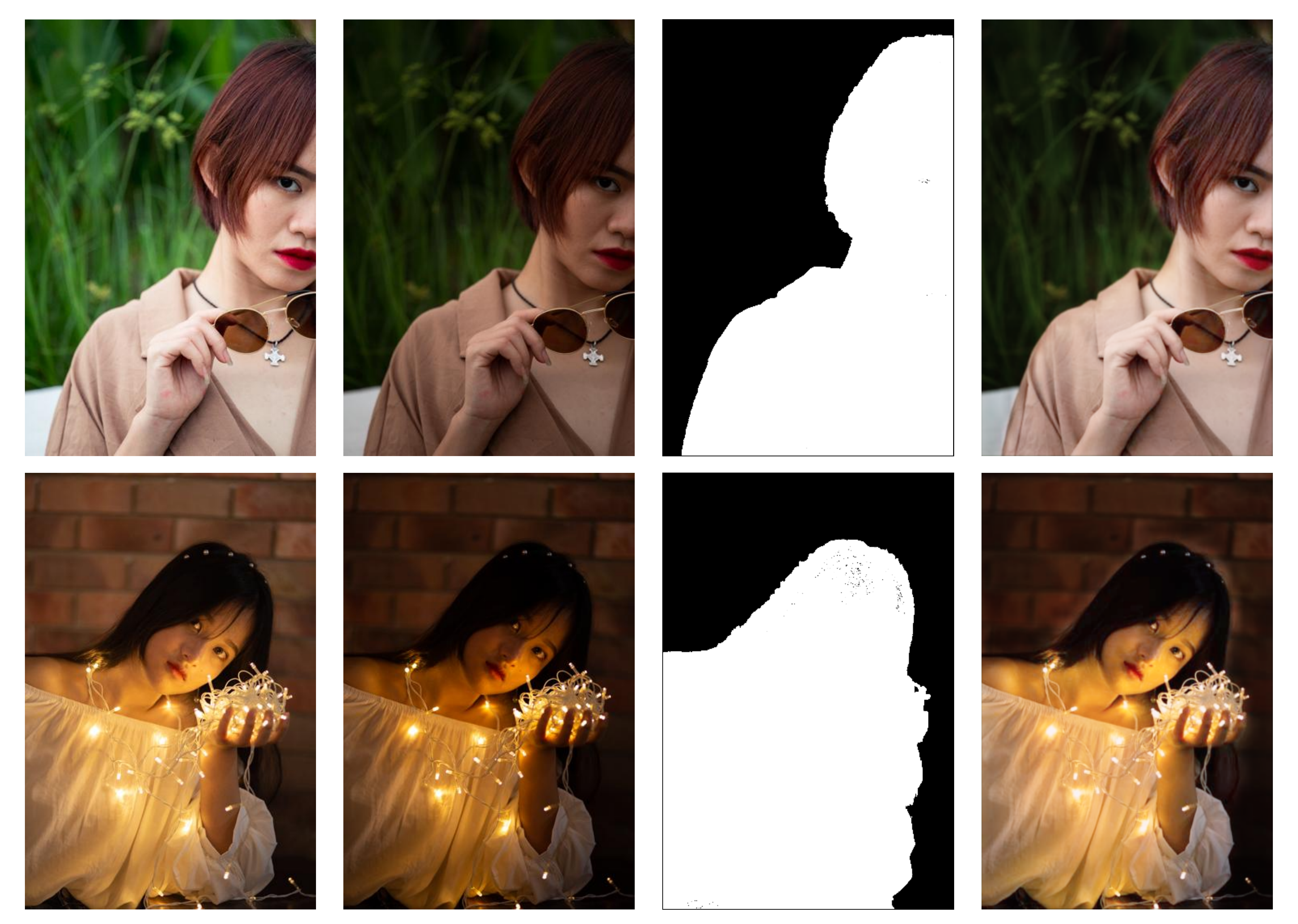}
        \caption{The performance of the STAR-DCE-\ourmethod model when given an low-light photo and a segmentation map as the image enhancement region. Left to right: ground truth image, original image, segmentation map, and output.}
        \label{PPR}
\end{figure}

We demonstrate how to achieve different degree of light based on user preference using DeepUPE.
In the DeepUPE model, we have the ability to manipulate the illumination map generated by the DeepUPE backbone. Similarly, in the STAR model, we can manipulate the predicted curve generated by the STAR backbone. Fig.~\ref{localllie-degree} illustrates some example results of the DeepUPE-\ourmethod model.

\subsection{Portrait Enhancement}
Our method can also be employed to achieve local LLIE with semantic segmentation masks, and one notable example is the portrait mask. PPR10K~\cite{jie2021PPR10K} is a portrait photo dataset, where each image pair includes a raw portrait photo captured with varying exposure, a segmentation map, and expert-retouched photos. We select some underexposed photos and expert-retouched photos as training pairs to train \ourmethod.
Regarding the mask, we utilize the original segmentation map as the second channel of the mask, and for the first channel, we apply image erosion to the segmentation map. In Fig.~\ref{PPR}, it is evident that \ourmethod can effectively handle different shapes of portrait masks and enhance specific portrait regions.

\section{Limitation and Future Work}
In this paper, we have proposed a practical framework to achieve local LLIE, which is extended from existing global LLIE methods without modification to their backbones.
Our approach can be seen as a baseline in the field of local LLIE for future methods. 
However, the employment of RANLEN will result in additional parameters, although it's not heavy.
In the future, we plan to extend our approach with more lightweight modules via parameterized models. Moreover, we will explore more spatial-adaptive manners to achieve better local LLIE effects.

\section{Conclusion}    
This paper introduces a new task setting Local LLIE, which is to enhance a low-light image within a given mask. We propose \ourmethod to convert a global image enhancement model to enable local image enhancement by encoding mask information in some of the normalization layers of the model.  Accordingly, we apply different loss components to different spatial areas, where the key is the novel gradient-smooth loss in Area $B$ to achieve smooth transition effects. Experiments have shown that \ourmethod is effective in the new local enhancement setting. We demonstrate some practical applications, including controlling the degree of the light and portrait enhancement enabled by \ourmethod.

\small
\bibliographystyle{ieeenat_fullname}
\bibliography{references}

\begin{thebibliography}{28}
\providecommand{\natexlab}[1]{#1}
\providecommand{\url}[1]{\texttt{#1}}
\expandafter\ifx\csname urlstyle\endcsname\relax
  \providecommand{\doi}[1]{doi: #1}\else
  \providecommand{\doi}{doi: \begingroup \urlstyle{rm}\Url}\fi

\bibitem[Ba et~al.(2016)Ba, Kiros, and Hinton]{layernorm}
Jimmy~Lei Ba, Jamie~Ryan Kiros, and Geoffrey~E. Hinton.
\newblock Layer normalization, 2016.

\bibitem[Bychkovsky et~al.(2011)Bychkovsky, Paris, Chan, and Durand]{fivek}
Vladimir Bychkovsky, Sylvain Paris, Eric Chan, and Fr{\'e}do Durand.
\newblock Learning photographic global tonal adjustment with a database of
  input / output image pairs.
\newblock In \emph{CVPR}, 2011.

\bibitem[Cai et~al.(2023)Cai, Bian, Lin, Wang, Timofte, and
  Zhang]{Cai_2023_ICCV}
Yuanhao Cai, Hao Bian, Jing Lin, Haoqian Wang, Radu Timofte, and Yulun Zhang.
\newblock Retinexformer: One-stage retinex-based transformer for low-light
  image enhancement.
\newblock In \emph{Proceedings of the IEEE/CVF International Conference on
  Computer Vision (ICCV)}, 2023.

\bibitem[Chen et~al.(2019)Chen, Chen, Do, and Koltun]{sid}
Chen Chen, Qifeng Chen, Minh Do, and Vladlen Koltun.
\newblock Seeing motion in the dark.
\newblock In \emph{ICCV}, 2019.

\bibitem[Chen et~al.(2021)Chen, Wang, Guo, Xu, Deng, Liu, Ma, Xu, Xu, and
  Gao]{ipt}
Hanting Chen, Yunhe Wang, Tianyu Guo, Chang Xu, Yiping Deng, Zhenhua Liu, Siwei
  Ma, Chunjing Xu, Chao Xu, and Wen Gao.
\newblock Pre-trained image processing transformer.
\newblock In \emph{CVPR}, 2021.

\bibitem[Guo et~al.(2020)Guo, Li, Guo, Loy, Hou, Kwong, and Cong]{Zero-DCE}
Chunle Guo, Chongyi Li, Jichang Guo, Chen~Change Loy, Junhui Hou, Sam Kwong,
  and Runmin Cong.
\newblock Zero-reference deep curve estimation for low-light image enhancement.
\newblock In \emph{CVPR}, 2020.

\bibitem[Guo et~al.(2017)Guo, Li, and Ling]{lime}
Xiaojie Guo, Yu Li, and Haibin Ling.
\newblock {LIME}: Low-light image enhancement via illumination map estimation.
\newblock \emph{IEEE TIP}, 2017.

\bibitem[Haralick et~al.(1987)Haralick, Sternberg, and Zhuang]{mor}
Robert~M. Haralick, Stanley~R. Sternberg, and Xinhua Zhuang.
\newblock Image analysis using mathematical morphology.
\newblock \emph{PAMI}, 1987.

\bibitem[Ioffe and Szegedy(2015)]{batchnorm}
Sergey Ioffe and Christian Szegedy.
\newblock Batch normalization: Accelerating deep network training by reducing
  internal covariate shift.
\newblock In \emph{ICML}, 2015.

\bibitem[Kolesnikov et~al.(2021)Kolesnikov, Dosovitskiy, Weissenborn, Heigold,
  Uszkoreit, Beyer, Minderer, Dehghani, Houlsby, Gelly, Unterthiner, and
  Zhai]{vit}
Alexander Kolesnikov, Alexey Dosovitskiy, Dirk Weissenborn, Georg Heigold,
  Jakob Uszkoreit, Lucas Beyer, Matthias Minderer, Mostafa Dehghani, Neil
  Houlsby, Sylvain Gelly, Thomas Unterthiner, and Xiaohua Zhai.
\newblock An image is worth 16x16 words: Transformers for image recognition at
  scale.
\newblock In \emph{ICLR}, 2021.

\bibitem[Kosugi and Yamasaki(2020)]{lr}
Satoshi Kosugi and Toshihiko Yamasaki.
\newblock Unpaired image enhancement featuring
  reinforcement-learning-controlled image editing software.
\newblock In \emph{AAAI}, 2020.

\bibitem[Lei et~al.(2022)Lei, Guo, Yang, Cui, Xie, and Huang]{Lei_2022_CVPR}
Biwen Lei, Xiefan Guo, Hongyu Yang, Miaomiao Cui, Xuansong Xie, and Di Huang.
\newblock Abpn: Adaptive blend pyramid network for real-time local retouching
  of ultra high-resolution photo.
\newblock In \emph{CVPR}, 2022.

\bibitem[Liang et~al.(2021)Liang, Zeng, Cui, Xie, and Zhang]{jie2021PPR10K}
Jie Liang, Hui Zeng, Miaomiao Cui, Xuansong Xie, and Lei Zhang.
\newblock Ppr10k: A large-scale portrait photo retouching dataset with
  human-region mask and group-level consistency.
\newblock In \emph{Proceedings of the IEEE Conference on Computer Vision and
  Pattern Recognition}, 2021.

\bibitem[Moran et~al.(2020)Moran, Marza, McDonagh, Parisot, and
  Slabaugh]{deeplpf}
Sean Moran, Pierre Marza, Steven McDonagh, Sarah Parisot, and Gregory Slabaugh.
\newblock {DeepLPF}: Deep local parametric filters for image enhancement.
\newblock In \emph{CVPR}, 2020.

\bibitem[Rother et~al.(2011)Rother, Kiefel, Zhang, Sch\"{o}lkopf, and
  Gehler]{smooth1}
Carsten Rother, Martin Kiefel, Lumin Zhang, Bernhard Sch\"{o}lkopf, and Peter
  Gehler.
\newblock Recovering intrinsic images with a global sparsity prior on
  reflectance.
\newblock In \emph{NeurIPS}, 2011.

\bibitem[Tian and Cohen(2017)]{tian2017}
Qi-Chong Tian and Laurent~D. Cohen.
\newblock Global and local contrast adaptive enhancement for non-uniform
  illumination color images.
\newblock In \emph{2017 IEEE International Conference on Computer Vision
  Workshops (ICCVW)}, 2017.

\bibitem[Ulyanov et~al.(2017)Ulyanov, Vedaldi, and Lempitsky]{instancenorm}
Dmitry Ulyanov, Andrea Vedaldi, and Victor Lempitsky.
\newblock Instance normalization: The missing ingredient for fast stylization,
  2017.

\bibitem[Vaswani et~al.(2017)Vaswani, Shazeer, Parmar, Uszkoreit, Jones, Gomez,
  Kaiser, and Polosukhin]{transformer}
Ashish Vaswani, Noam Shazeer, Niki Parmar, Jakob Uszkoreit, Llion Jones,
  Aidan~N Gomez, {\L}ukasz Kaiser, and Illia Polosukhin.
\newblock Attention is all you need.
\newblock In \emph{NeurIPS}, 2017.

\bibitem[Wang et~al.(2019)Wang, Zhang, Fu, Shen, Zheng, and Jia]{DeepUPE}
Ruixing Wang, Qing Zhang, Chi-Wing Fu, Xiaoyong Shen, Wei-Shi Zheng, and Jiaya
  Jia.
\newblock Underexposed photo enhancement using deep illumination estimation.
\newblock In \emph{CVPR}, 2019.

\bibitem[Wang et~al.(2021)Wang, Li, Peng, Ma, Wang, Song, and Yan]{3dlut2}
Tao Wang, Yong Li, Jingyang Peng, Yipeng Ma, Xian Wang, Fenglong Song, and
  Youliang Yan.
\newblock Real-time image enhancer via learnable spatial-aware 3d lookup
  tables.
\newblock In \emph{Int. Conf. Comput. Vis.}, 2021.

\bibitem[Wang et~al.(2023)Wang, Yu, Yang, Guo, Chau, Kot, and
  Wen]{wang2023exposurediffusion}
Yufei Wang, Yi Yu, Wenhan Yang, Lanqing Guo, Lap-Pui Chau, Alex~C Kot, and
  Bihan Wen.
\newblock Exposurediffusion: Learning to expose for low-light image
  enhancement.
\newblock \emph{arXiv preprint arXiv:2307.07710}, 2023.

\bibitem[Wang et~al.(2004)Wang, Bovik, Sheikh, and Simoncelli]{ssim}
Zhou Wang, A.C. Bovik, H.R. Sheikh, and E.P. Simoncelli.
\newblock Image quality assessment: from error visibility to structural
  similarity.
\newblock \emph{IEEE TIP}, 2004.

\bibitem[Wu et~al.(2022)Wu, Weng, Zhang, Wang, Yang, and Jiang]{uretinexnet}
Wenhui Wu, Jian Weng, Pingping Zhang, Xu Wang, Wenhan Yang, and Jianmin Jiang.
\newblock Uretinex-net: Retinex-based deep unfolding network for low-light
  image enhancement.
\newblock In \emph{2022 IEEE/CVF Conference on Computer Vision and Pattern
  Recognition (CVPR)}, 2022.

\bibitem[Yi et~al.(2023)Yi, Xu, Zhang, Tang, and Ma]{diff-retinex}
Xunpeng Yi, Han Xu, Hao Zhang, Linfeng Tang, and Jiayi Ma.
\newblock Diff-retinex: Rethinking low-light image enhancement with a
  generative diffusion model.
\newblock In \emph{Proceedings of the IEEE/CVF International Conference on
  Computer Vision (ICCV)}, 2023.

\bibitem[Yu et~al.(2018)Yu, Liu, Zhang, Qu, Zhao, and Zhang]{deepexposure}
Runsheng Yu, Wenyu Liu, Yasen Zhang, Zhi Qu, Deli Zhao, and Bo Zhang.
\newblock {DeepExposure}: Learning to expose photos with asynchronously
  reinforced adversarial learning.
\newblock In \emph{NeurIPS}, 2018.

\bibitem[Zamir et~al.(2020)Zamir, Arora, Khan, Hayat, Khan, Yang, and
  Shao]{2020MIRNet}
Syed~Waqas Zamir, Aditya Arora, Salman Khan, Munawar Hayat, Fahad~Shahbaz Khan,
  Ming-Hsuan Yang, and Ling Shao.
\newblock Learning enriched features for real image restoration and
  enhancement.
\newblock In \emph{Eur. Conf. Comput. Vis.}, 2020.

\bibitem[Zeng et~al.(2020)Zeng, Cai, Li, Cao, and Zhang]{3dlut1}
Hui Zeng, Jianrui Cai, Lida Li, Zisheng Cao, and Lei Zhang.
\newblock Learning image-adaptive 3d lookup tables for high performance photo
  enhancement in real-time.
\newblock \emph{IEEE Transactions on Pattern Analysis and Machine
  Intelligence}, 2020.

\bibitem[Zhang et~al.(2021)Zhang, Jiang, Jiang, Wang, Luo, and Gu]{star}
Zhaoyang Zhang, Yitong Jiang, Jun Jiang, Xiaogang Wang, Ping Luo, and Jinwei
  Gu.
\newblock Star: A structure-aware lightweight transformer for real-time image
  enhancement.
\newblock In \emph{ICCV}, 2021.

\end{thebibliography}

\end{document}